\documentclass[10pt,twocolumn,letterpaper]{article}
\usepackage{cvpr}
\usepackage[table,xcdraw]{xcolor}
\usepackage[accsupp]{axessibility}

\definecolor{cvprblue}{rgb}{0.21,0.49,0.74}
\usepackage[pagebackref,breaklinks,colorlinks,allcolors=cvprblue]{hyperref}
\usepackage{multirow}
\usepackage{amsmath,amssymb,amsfonts}  
\usepackage{graphicx}  
\usepackage{bbding}
\usepackage{booktabs}
\usepackage{nicematrix}
\usepackage{makecell}
\usepackage{pifont}
\usepackage[accsupp]{axessibility}

\newcommand{\greenredmark}{{\color{green}{\Checkmark}}\kern-1.2ex\raisebox{1ex}{\color{red}{\rotatebox[origin=c]{125}{\textbf{--}}}}}

\definecolor{cjh}{HTML}{e61849}
\definecolor{zzl}{HTML}{ee9010}
\definecolor{mygreen}{HTML}{d8f1cf}
\definecolor{mycolor}{HTML}{e9e9e9}
\def\paperID{1048} 
\def\confName{CVPR}
\def\confYear{2026}
\title{Seeing as Experts Do: A Knowledge-Augmented Agent for \\ Open-Set {Fine-Grained Visual Understanding}}
\author{Junhan Chen\textsuperscript{†}, Zilu Zhou\textsuperscript{†}, Yujun Tong\textsuperscript{†}, Dongliang Chang*, Yitao Luo, and Zhanyu Ma\\
{\small School of Artificial Intelligence, Beijing University of Posts and Telecommunications, China} \\
{\tt\small\{chenjunhan, zhouzilu2019, tongyujun, changdongliang, lyt2022, mazhanyu\}@bupt.edu.cn}
}

\makeatletter
\renewenvironment{abstract}{
  \centerline{\large\bf Abstract}
  \vspace{0.5\baselineskip}
  \itshape\noindent\ignorespaces
}{\par\unskip}
\makeatother

\begin{document}
\twocolumn[{
\renewcommand\twocolumn[1][]{#1}%
\maketitle
\vspace{-2.5em}
\begin{center}
    \centering
    \captionsetup{type=figure}
    \includegraphics[width=0.95\textwidth]{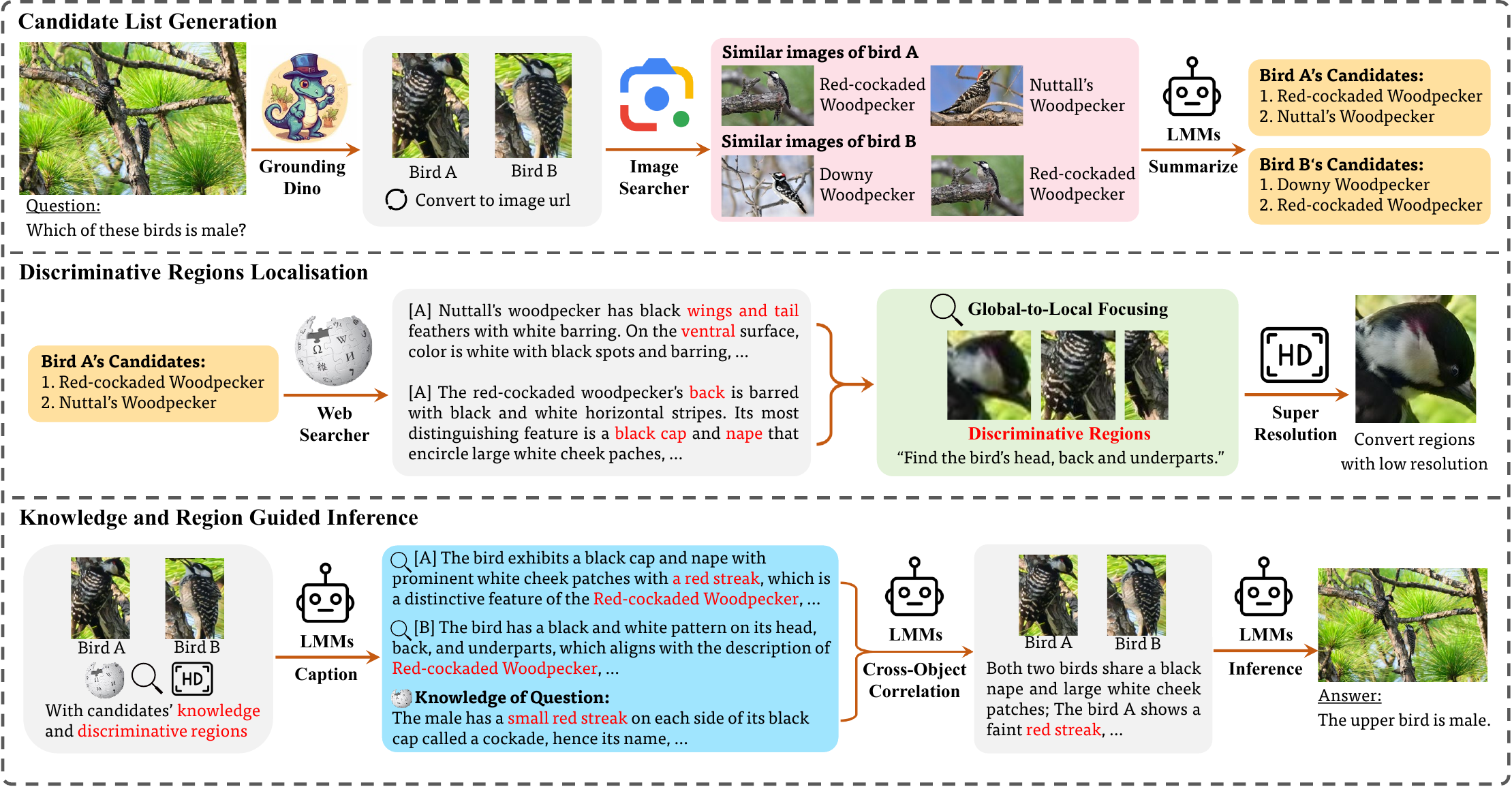}
    \captionof{figure}{{\textbf{Overview of the Knowledge-Augmented Fine-Grained Reasoning Agent (KFRA).}} The agent operates through a three-stage closed reasoning loop that emulates expert analysis. (1) \textit{Candidate List Generation}: KFRA performs open-vocabulary detection and web-scale retrieval to propose category hypotheses. (2) \textit{Discriminative Regions Localisation}: For each hypothesis, it retrieves relevant textual knowledge and aligns distinctive cues such as colour, structure, or behaviour with corresponding visual regions through a global-to-local focusing mechanism. (3) \textit{Knowledge and Region Guided Inference}: The agent integrates all multimodal evidence to perform factual and interpretable reasoning across objects and tasks.}
    \label{fig:figure1}
\end{center}%
}]

\renewcommand{\thefootnote}{}
\footnotetext{
\hspace{-1.5em}
\raggedright
\textsuperscript{†} Equal contributions. * Corresponding author.
}

\begin{abstract}
Fine-grained visual understanding is shifting from static classification to knowledge-augmented reasoning, where models must justify as well as recognise. Existing approaches remain limited by closed-set taxonomies and single-label prediction, leading to significant degradation under open-set or context-dependent conditions.
We present the Knowledge-Augmented Fine-Grained Reasoning Agent (KFRA), a unified framework that transforms fine-grained perception into evidence-driven reasoning. KFRA operates through a three-stage closed reasoning loop that emulates expert analysis. It first performs open-vocabulary detection and web-scale retrieval to generate category hypotheses. It then conducts discriminative regions localisation by aligning textual knowledge with visual evidence through a global-to-local focusing mechanism. Finally, it integrates all multimodal evidence within a large multimodal model to perform interpretable reasoning.
Unlike existing agents that treat retrieval and reasoning as independent processes, KFRA establishes a retrieval–grounding coupling that converts retrieved knowledge into spatially grounded evidence for verification. This design enables factual, interpretable, and task-agnostic reasoning across diverse fine-grained scenarios. To evaluate this capability, we construct FGExpertBench, a benchmark designed to assess reasoning depth and cross-task generalisation across six knowledge dimensions. Extensive experiments demonstrate that KFRA consistently surpasses both standalone large multimodal models and current agent frameworks, achieving up to 19 percent improvement in reasoning accuracy and delivering evidence-grounded interpretability in open-set fine-grained visual understanding.
\end{abstract}

\section{Introduction}
Fine-grained visual understanding is no longer about naming what an object is, but about explaining why it is what it is. The challenge is not recognition but reasoning, connecting perceptual evidence to the structured knowledge that defines expert understanding~\cite{wei2021fine}. Despite years of progress, most fine-grained systems still behave like refined classifiers~\cite{du2023fly, yang2025adaptive, wang2024multi}, accurate within a closed taxonomy but incapable of reasoning beyond it.

\begin{figure}[t]
\centering
\includegraphics[width=0.95\linewidth]{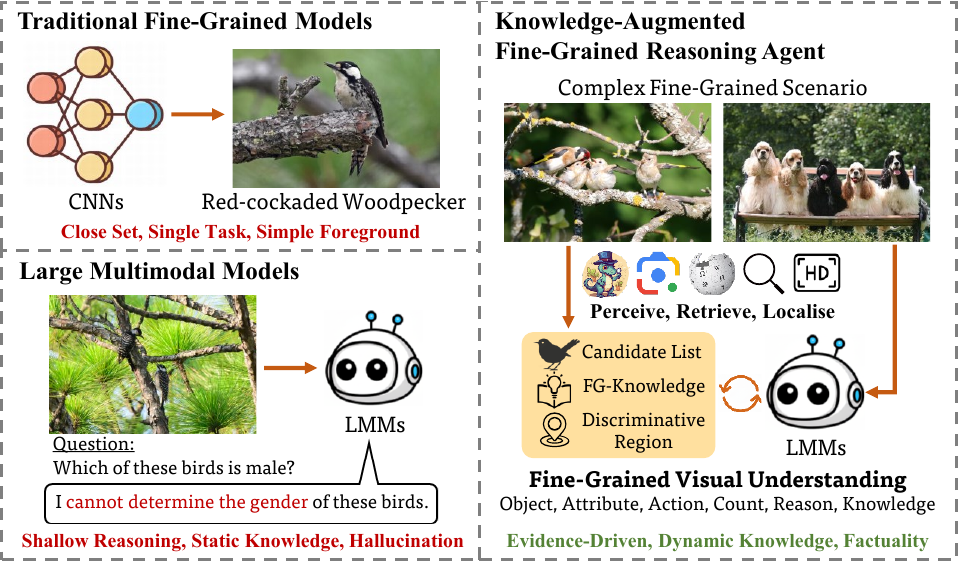}
\caption{\textbf{Comparison of Fine-Grained Paradigms.} 
KFRA advances fine-grained understanding from static classification to evidence-driven reasoning, leveraging dynamic knowledge and retrieval–grounding coupling to achieve factual, interpretable, and task-agnostic understanding.}
\vspace{-1.5em}
\label{fig:figure2}
\end{figure}

Recent advances in transformer architectures~\cite{huang2025attention, bi2025universal}, {large multimodal models post-training}~\cite{li2025dyfo, he2025analyzing, wang2025cinetechbench}, and retrieval-augmented generation~\cite{ru2024ragchecker, han2025fine} have extended perception to open-vocabulary settings. Yet, the optimisation goal remains unchanged: predict a single label from a fixed taxonomy {(as shown in Figure~\ref{fig:figure2})}. External cues such as attributes, hierarchies, or prompts enrich semantics but do not alter this objective~\cite{chang2021your,chang2023making}. Compressing a complex instance into one categorical token collapses expert knowledge into a flat decision boundary and leaves no capacity for reasoning about unseen subtypes, abnormal states, or contextual variations. In practice, existing fine-grained models lose up to 30–40\% accuracy~\cite{chang2023erudite} when confronted with unseen species or domains, exposing the fragility of closed-set formulations. This motivates a paradigm shift from label prediction to evidence-driven reasoning.

Fine-grained understanding is evolving from static classification to knowledge-grounded reasoning, where models must justify as well as recognise. Human experts, when faced with ambiguous or novel instances, do not rely on memory alone—they hypothesise candidate categories, retrieve references, focus on discriminative traits, and verify these hypotheses against factual knowledge before concluding. This iterative process of observing, hypothesising, and verifying forms a chain of evidence that connects perception with external knowledge. Reproducing this process computationally requires a system that can perceive, retrieve, localise, and reason across multiple tasks rather than a single classification objective. Existing approaches fall short of constructing such reasoning chains. To bridge this gap, we introduce the {Knowledge-Augmented Fine-Grained Reasoning Agent (KFRA)}, which reformulates fine-grained understanding as an open-set reasoning process grounded in open-world knowledge (as shown in Figure~\ref{fig:figure1}).

KFRA generalises beyond recognition to a spectrum of fine-grained reasoning tasks, including object identification, attribute comparison, action understanding, counting, causal analysis, and knowledge inference. It operates through a three-stage closed reasoning loop that emulates expert analysis:
(1) Candidate List Generation: KFRA performs open-vocabulary detection to isolate visual entities and conducts web-scale image retrieval to gather visually similar examples, forming candidate hypotheses.
(2) Discriminative Regions Localisation: For each hypothesis, it retrieves relevant textual knowledge describing distinctive cues such as colour, structure, or behaviour, and aligns these descriptions with image regions through a global-to-local focusing mechanism. When critical details are missing, an automatic detail-enhancement tool refines local regions for higher discriminability.
(3) Knowledge and Region Guided Inference: KFRA integrates all available evidence, including hypotheses, textual knowledge, and grounded regions, within a large multimodal model to reason across objects and modalities and to produce interpretable conclusions.
This closed-loop design ensures coherence between retrieval, localisation, and inference, allowing KFRA to adapt across fine-grained tasks without retraining.

Unlike general-purpose agents~\cite{wu2025mmsearch,song2025beyond, lai2025mini} that treat retrieval and inference as loosely connected steps, KFRA establishes a retrieval–grounding coupling that transforms large multimodal models from passive label predictors into active evidence builders. Retrieved knowledge is not auxiliary context but an actionable signal that guides spatial grounding and hypothesis verification. This coupling enables factual, spatially-grounded, and task-agnostic reasoning across diverse fine-grained scenarios.

\begin{figure*}[t]
\centering
\includegraphics[width=0.97\linewidth]{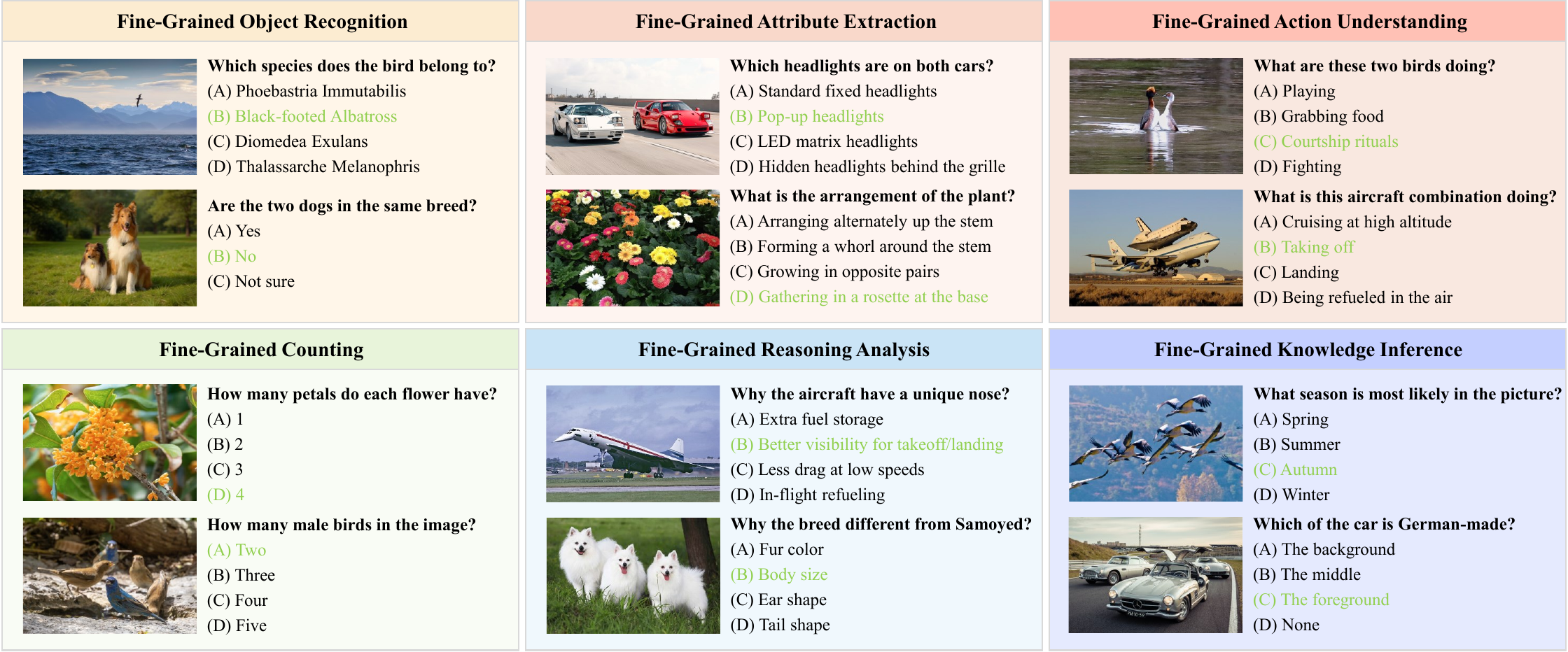}
\caption{\textbf{Overview of \textit{FGExpertBench}.} The benchmark introduces a structured taxonomy of fine-grained visual understanding tasks that span multiple perception and reasoning dimensions. Each category includes representative examples illustrating its task type, while the overall design enables systematic evaluation of large multimodal models in terms of reasoning depth, cross-task generalisation, and evidence grounding.}
\vspace{-1.6em}
\label{fig:figure3}
\end{figure*}

To quantify these capabilities, we construct \textit{FGExpertBench}, a benchmark that measures reasoning depth and cross-task generalisation rather than recognition accuracy. It comprises 300 images and 1,500 question–answer pairs across six knowledge dimensions: object recognition, attribute extraction, action understanding, counting, reasoning analysis, and knowledge inference (Figure~\ref{fig:figure3}). Each task requires synthesising perceptual cues with external knowledge to produce a justified answer. Experiments show that KFRA consistently surpasses both standalone large multimodal models and existing agentic frameworks. When integrated with strong backbones such as GLM-4.5V~\cite{team2025glm} or Qwen3-VL~\cite{yang2025qwen3}, KFRA improves reasoning accuracy by up to 19\% and delivers evidence-grounded interpretability across all dimensions.

In summary, this work makes three key contributions:
(1) It introduces KFRA, a knowledge-augmented reasoning framework that unifies diverse fine-grained tasks under a single evidence-driven paradigm.
(2) It develops a closed-loop pipeline that couples retrieval, grounded localisation, and multimodal inference, enabling zero-shot reasoning across tasks and domains.
(3) It releases \textit{FGExpertBench}, a benchmark for evaluating reasoning depth and cross-task generalisation in fine-grained understanding.
Together, these advances move fine-grained vision beyond static taxonomy recognition toward dynamic, knowledge-grounded reasoning—bringing machine perception one step closer to expert-level cognition.

\section{Related Work}
Understanding how fine-grained reasoning emerges from perception and knowledge has long been a central challenge in visual understanding. Existing research can be grouped into three threads that our work builds upon and extends.

\noindent\textbf{Fine-Grained Visual Understanding.}
Early work on fine-grained recognition~\cite{chen2025fcnet,du2020fine} largely treated the problem as classification within a fixed taxonomy, focusing on distinguishing visually similar subcategories through local part detection or attention-based feature aggregation. While effective on benchmarks, such models relied heavily on category-specific supervision and lacked interpretability beyond label prediction.
Recent research has begun to move toward richer understanding through hierarchical reasoning~\cite{liang2023hierarchical, liang2024learning}, multi-view perception~\cite{du2023multi, yin2022duplex}, and visual–semantic alignment~\cite{chang2022coder,weng2023cad}. These approaches introduce attributes, hierarchies, or textual cues to enhance semantics, yet the optimisation target remains unchanged: predict a single label. As a result, they struggle to reason about unseen subtypes or contextual variations, revealing the limits of closed-set paradigms.
Our work departs from these data-driven formulations by reformulating fine-grained understanding as an open-set reasoning process that constructs an explicit chain of evidence connecting perception and knowledge.

\begin{figure*}[t]
\centering
\includegraphics[width=0.98\linewidth]{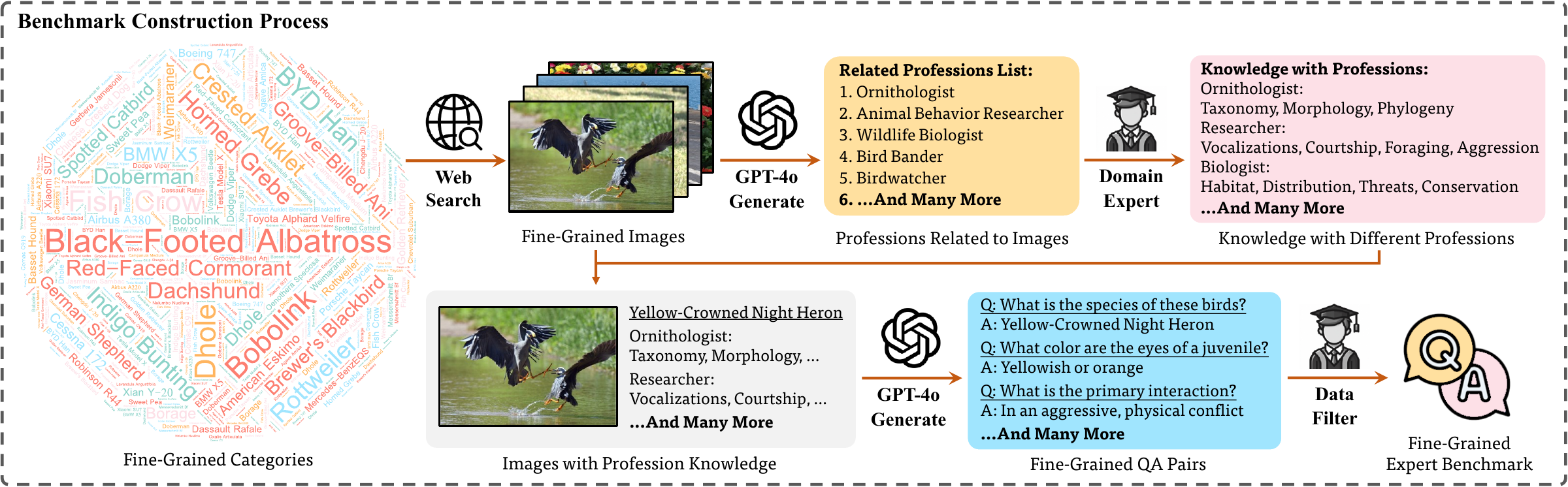}
\caption{\textbf{The construction pipeline of \textit{FGExpertBench}.} We first collect diverse fine-grained categories followed by retrieving corresponding images through web search. GPT-4o is then used to identify potential professional communities relevant to the image content, and domain experts check these professions while providing the corresponding knowledge scopes they focus on. Based on the curated knowledge, GPT-4o generates domain-oriented question-answer pairs. After data filtering and quality control, the finalized benchmark is produced for evaluating large multimodal models on fine-grained visual understanding.} 
\vspace{-1em}
\label{fig:figure4}
\end{figure*}

\noindent\textbf{Large Multimodal Models.}
Large multimodal models (LMMs)~\cite{huang2026step3,li2025llava} have reshaped visual understanding by jointly learning vision–language representations through large-scale pre-training. Modern LMMs such as BLIP~\cite{li2022blip}, LLaVA~\cite{liu2023visual}, and Qwen3-VL~\cite{yang2025qwen3} integrate visual encoders and cross-modal attention to perform open-vocabulary recognition~\cite{liu2024grounding}, instruction following~\cite{liu2023visual}, and compositional reasoning~\cite{ju2024brushnet}. However, most LMMs are designed for perception-oriented tasks and lack mechanisms for evidence construction or knowledge-grounded verification. Their predictions remain pattern-based rather than reason-based, limiting generalisation in fine-grained or domain-shifted settings.
KFRA leverages LMMs not as static recognisers but as active reasoning engines that interact with external knowledge sources. Through dynamic retrieval and grounded verification, it enables factual and interpretable reasoning beyond the capacity of monolithic LMMs.

\noindent\textbf{Web and Knowledge-Augmented Agents.}
With the rapid progress of large language models in tool use, autonomous agents have become a promising paradigm for external knowledge acquisition. Early agents~\cite{liu2023webglm,nakano2022webgpt} were limited to text-only question answering, while later systems~\cite{wu2025mmsearch,zhang2024vision,jiang2024mmsearch} introduced multimodal search and reinforcement learning to perform multi-step reasoning. Nevertheless, most existing agents treat retrieval and inference as independent stages and fail to ground retrieved information in visual evidence.
Our work advances this line by introducing a closed reasoning loop that couples retrieval with grounded verification. Unlike prior agents that retrieve knowledge passively, KFRA integrates web-scale search, textual grounding, and multimodal inference into a unified process. This retrieval–grounding coupling transforms fine-grained reasoning from label prediction into evidence construction, enabling factual, interpretable, and task-agnostic understanding across diverse visual domains.

\section{Methodology}
We formulate fine-grained understanding as a \textbf{closed reasoning loop} that connects perception, knowledge, and inference. 
Given an input image $X$ and a natural-language question $q$, the proposed {Knowledge-Augmented Fine-Grained Reasoning Agent (KFRA)} performs three sequential stages: \textit{candidate list generation}, \textit{discriminative regions localisation}, and \textit{knowledge and region guided inference}, which together emulate the analytical process of human experts. 
To enable systematic evaluation, we also introduce a new benchmark, \textit{FGExpertBench}, as described in Section~\ref{subsec:bench}.

\subsection{Benchmark Construction}
\label{subsec:bench}
\textit{FGExpertBench} is designed to evaluate the capability of models to bridge perceptual understanding with domain-specific knowledge. 
Unlike conventional visual question answering datasets, it adopts a \textbf{category-based reasoning} setting, where correct answers depend on recognising the depicted instance and reasoning with its corresponding expert knowledge. 
The benchmark is constructed through a semi-automated pipeline combining open-set data sampling, model-assisted generation, and expert verification.

\begin{table}[t]
\centering
\resizebox{0.4\textwidth}{!}{
\fontsize{8pt}{10pt}\selectfont
\setlength{\tabcolsep}{7pt}
\begin{NiceTabular}{l|c|c|c|c|c}
\toprule
\makecell[l]{\textbf{Dimensions}} &
\textit{\makecell[c]{FOCI \\[-2pt] Bench}} &
\textit{\makecell[c]{FG \\[-2pt] BMK}} &
\textit{\makecell[c]{KVG \\[-2pt] Bench}} &
\textit{\makecell[c]{Naut \\[-2pt] Data}} &
\textit{\makecell[c]{FGExpert \\[-2pt] Bench}} \\
\addlinespace[1pt]
\midrule
Object & \Block[fill=mygreen]{1-1}{\textcolor{black!80}{\ding{51}}} & \Block[fill=mygreen]{1-1}{\textcolor{black!80}{\ding{51}}} & \Block[fill=mygreen]{1-1}{\textcolor{black!80}{\ding{51}}} & \Block[fill=mygreen]{1-1}{\textcolor{black!80}{\ding{51}}} & \Block[fill=mygreen]{1-1}{\textcolor{black!80}{\ding{51}}} \\
Attribute & \textcolor{black!80}{\ding{55}} & \Block[fill=mygreen]{1-1}{\textcolor{black!80}{\ding{51}}} & \textcolor{black!80}{\ding{55}} & \textcolor{black!80}{\ding{55}} & \Block[fill=mygreen]{1-1}{\textcolor{black!80}{\ding{51}}} \\
Action & \textcolor{black!80}{\ding{55}} & \textcolor{black!80}{\ding{55}} & \textcolor{black!80}{\ding{55}} & \textcolor{black!80}{\ding{55}} & \Block[fill=mygreen]{1-1}{\textcolor{black!80}{\ding{51}}} \\
Count & \textcolor{black!80}{\ding{55}} & \textcolor{black!80}{\ding{55}} & \textcolor{black!80}{\ding{55}} & \Block[fill=mygreen]{1-1}{
\textcolor{black!80}{\ding{51}}} & \Block[fill=mygreen]{1-1}{\textcolor{black!80}{\ding{51}}} \\
Reason & \textcolor{black!80}{\ding{55}} & \textcolor{black!80}{\ding{55}} & \Block[fill=mygreen]{1-1}{\textcolor{black!80}{\ding{51}}} & \Block[fill=mygreen]{1-1}{\textcolor{black!80}{\ding{51}}} & \Block[fill=mygreen]{1-1}{\textcolor{black!80}{\ding{51}}} \\
Knowledge & \textcolor{black!80}{\ding{55}} & \textcolor{black!80}{\ding{55}} & \textcolor{black!80}{\ding{55}} & \textcolor{black!80}{\ding{55}} & \Block[fill=mygreen]{1-1}{\textcolor{black!80}{\ding{51}}} \\
\bottomrule
\end{NiceTabular}
}
\caption{\textbf{Comparison with Similar Benchmarks.} 
\textit{FGExpertBench} provides broader and more comprehensive coverage of fine-grained understanding dimensions than existing benchmarks.}
\vspace{-1.5em}
\label{tab:table1}
\end{table}

\noindent\textbf{Benchmark Dimensions and Taxonomy.}
The benchmark covers six reasoning dimensions that reflect the progressive nature of fine-grained understanding: 
object recognition, attribute extraction, action understanding, counting, reasoning analysis, and knowledge inference. 
Each dimension evaluates a distinct aspect of perception and reasoning, as illustrated in Figure~\ref{fig:figure3}.

\begin{itemize}
    \item \textbf{Object Recognition:} Identify the object’s fine-grained category or determine whether instances belong to the same superclass. 
    \item \textbf{Attribute Extraction:} Perceive discriminative attributes (e.g., colour, texture, morphology) and reason about their semantic implications. 
    \item \textbf{Action Understanding:} Interpret subject behaviours and infer purpose (e.g., feeding, courtship). 
    \item \textbf{Counting:} Identify and enumerate fine-grained visual or conceptual entities. 
    \item \textbf{Reasoning Analysis:} Explain causal or functional relationships underlying visual phenomena. 
    \item \textbf{Knowledge Inference:} Integrate external knowledge for higher-level reasoning, such as environment, geography, or culture.
\end{itemize}

We analyse the coverage of reasoning dimensions across existing fine-grained benchmarks, as shown in Table~\ref{tab:table1}. 
The comparison reveals a clear gap. 
Current datasets such as \textit{FOCI-Bench}~\cite{geigle2024african}, \textit{FG-BMK}~\cite{yu2025benchmarking}, \textit{KVG-Bench}~\cite{ma2025deepperception}, and \textit{NautData}~\cite{xu2025nautilus} focus mainly on object recognition, with limited support for attributes, actions, or reasoning. 
\textit{FGExpertBench} is designed to fill this gap by providing balanced coverage across all dimensions and enabling a more systematic evaluation of fine-grained reasoning.

\noindent\textbf{Benchmark Construction Pipeline.}
As shown in Figure~\ref{fig:figure4}, we first assemble a broad category space from both traditional fine-grained datasets 
(e.g., CUB-200-2011~\cite{wah2011caltech}, FGVC-Aircraft~\cite{maji2013fine}, Stanford Dogs~\cite{khosla2011novel}) and open-world web sources to ensure diverse coverage. 
Unlike conventional datasets featuring single-object scenes, \textit{FGExpertBench} deliberately includes multi-object and complex contexts to evaluate reasoning robustness.

For each category, web retrieval is performed, followed by automatic semantic analysis using GPT-4o to identify related professional communities (e.g., ornithologists, botanists) and their focus dimensions. 
Domain experts then refine this mapping and guide GPT-4o to generate accurate question–answer pairs based on verified domain knowledge. 
All samples are further reviewed to ensure factual accuracy and conceptual diversity.

This process reduces annotation effort while maintaining expert-level quality. 
The final benchmark contains 300 images and 1,500 QA pairs distributed across six reasoning dimensions, offering a balanced and challenging testbed for knowledge-grounded reasoning. The statistics of the benchmark are provided in \textit{Supplementary Material}.

\subsection{Model Architecture}
\label{subsec:model}
KFRA operationalises expert reasoning through a \textbf{three-stage closed reasoning loop}, where each stage progressively refines and verifies the previous one.  
This architecture redefines fine-grained understanding as an evidence-driven reasoning process that connects perception, retrieval, and knowledge grounding within a unified framework (as shown in Figure~\ref{fig:figure1}). 
The agent is implemented through a set of specialised perception and reasoning tools coordinated by a large multimodal controller, whose implementation details are provided in Section~\ref{subsec:exp_set}.

\noindent\textbf{Stage 1: Candidate List Generation.}
The first stage constructs an open-set hypothesis space guided by visual–semantic retrieval.
Given an input image $X$ and question $q$, KFRA first performs open-vocabulary detection via $\mathcal{F}_{det}(\cdot)$:
\begin{equation}
\{x_i\}_{i=1}^{N} = \mathcal{F}_{det}(X),
\end{equation}
where $x_i$ denotes the $i$-th detected entity (i.e., an object region), and $N$ is the total number of detected objects in $X$.  
For each $x_i$, the image retriever $\mathcal{S}_{img}(\cdot)$ searches the web for visually similar examples and associated captions:
\begin{equation}
\mathcal{R}^{\text{img}}_i = \{(I_{ij}, T_{ij})\}_{j=1}^{M},
\end{equation}
where $I_{ij}$ and $T_{ij}$ denote the $j$-th retrieved image and its accompanying textual description, respectively, and $M$ is the number of top-ranked retrieval results retained as visual evidence.
A large multimodal model $\mathcal{F}_{lmm}(\cdot)$ then integrates these results with the query $q$ to generate a ranked list of category hypotheses:
\begin{equation}
C_i = \{(c, p_i(c)) \mid c \in \mathcal{Y}_i\},
\end{equation}
where $\mathcal{Y}_i$ represents the open-set label space inferred from the retrieved content, $c$ is a candidate category, and $p_i(c)$ denotes the confidence assigned to hypothesis $c$ by $\mathcal{F}_{lmm}(\cdot)$.  
This stage departs from conventional closed-set classification by constructing a retrieval-augmented hypothesis space, enabling open-world generalisation without dataset-specific supervision.

\noindent\textbf{Stage 2: Discriminative Regions Localisation.}
The second stage grounds retrieved knowledge to spatial evidence through knowledge-guided localisation.
For each candidate hypothesis $c \in C_i$, the textual retriever $\mathcal{S}_{tex}(\cdot)$ extracts relevant factual descriptions:
\begin{equation}
\mathcal{K}_{i,c} = \mathcal{S}_{tex}(c),
\end{equation}
where $\mathcal{K}_{i,c}$ is a set of textual knowledge related to category $c$, typically describing morphological, behavioural, or contextual cues (e.g., “red beak”, “striped wings”). 
The textual knowledge $\mathcal{K}_{i,c}$ is then parsed into a structured set of discriminative attributes:
\begin{equation}
\mathcal{A}_{i,c} = \{a^{(k)}_{i,c}\}_{k=1}^{m_{i,c}},
\end{equation}
where each $a^{(k)}_{i,c}$ denotes the $k$-th discriminative cue (such as a body part or distinctive colour) and $m_{i,c}$ is the number of cues extracted from $\mathcal{K}_{i,c}$.  
A global-to-local focusing module $\mathcal{F}_{foc}(\cdot)$ then aligns each cue with its corresponding visual region:
\begin{equation}
(\mathcal{M}^{(k)}_{i,c}, s^{(k)}_{i,c}) = \mathcal{F}_{foc}(x_i, a^{(k)}_{i,c}),
\end{equation}
where $\mathcal{M}^{(k)}_{i,c}\in [0,1]^{h\times w}$ is the attention mask indicating the spatial distribution of cue $a^{(k)}_{i,c}$ on region $x_i$, and $s^{(k)}_{i,c}$ is the alignment confidence.
The focusing module operates in a coarse-to-fine manner: the global stage employs CLIP-style semantic similarity to roughly locate relevant regions, while the local stage applies patch-level attention refinement to delineate precise boundaries.  
When fine details are missing or misaligned (i.e., $\max_k s^{(k)}_{i,c} < \tau$), the OseDiff super-resolution enhancer $\mathcal{F}_{sr}(\cdot)$ reconstructs the most confident region to recover high-frequency details:
\begin{equation}
\tilde{x}_i = \mathcal{F}_{sr}(x_i \odot \mathcal{M}^{(k^\star)}_{i,c}), \quad k^\star = \arg\max_k s^{(k)}_{i,c},
\end{equation}
where $\tilde{x}_i$ is the enhanced region centred on the most reliable cue $a^{(k^\star)}_{i,c}$, and $\odot$ denotes element-wise masking.
The enhanced patch is reprocessed by $\mathcal{F}_{foc}(\cdot)$ to refine localisation.
Through this stage, KFRA realises a retrieval–grounding coupling mechanism: textual knowledge directs attention localisation, while visual evidence iteratively refines retrieved cues, forming a bidirectional link between semantics and perception.

\noindent\textbf{Stage 3: Knowledge and Region Guided Inference.}
The final stage integrates multimodal evidence into a unified reasoning process.
For each object, an evidence tuple is constructed:
\begin{equation}
E_i = \{(c, p_i(c), \mathcal{K}_{i,c}, \mathcal{A}_{i,c}, \mathcal{M}_{i,c}) \mid c \in C_i\},
\end{equation}
where $E_i$ aggregates all available information for the $i$-th object, including category hypotheses $c$, their confidence scores $p_i(c)$, associated textual knowledge $\mathcal{K}_{i,c}$, discriminative attributes $\mathcal{A}_{i,c}$, and grounded visual masks $\mathcal{M}_{i,c}$.
The large multimodal model then performs reasoning conditioned on the accumulated evidence:
\begin{equation}
P(y|X,q) = \mathcal{F}_{lmm}(X, q, \{E_i\}_{i=1}^{N}),
\end{equation}
\vspace{-1em}
\begin{equation}
\hat{y} = \arg\max_y P(y|X,q),
\end{equation}
where $P(y|X,q)$ represents the probability distribution over possible answers given image $X$ and question $q$, and $\hat{y}$ is the final predicted answer.
If $\hat{y}$ exhibits low confidence, the controller re-invokes earlier stages to refine hypotheses or localisation, completing the self-corrective reasoning loop.  
Through this design, KFRA transforms large multimodal models from passive recognisers into active reasoning agents capable of constructing, verifying, and revising evidence dynamically.

\noindent\textbf{Discussion.}
KFRA unifies perception, retrieval, and reasoning within a self-corrective framework, achieving factual, interpretable, and task-agnostic fine-grained understanding across open-world scenarios without any task-specific retraining.

\section{Experiments}
In this section, we conduct extensive experiments to evaluate and analyze our proposed method. Section~\ref{subsec:exp_set} describes experimental settings, Section~\ref{subsec:exp_compare} compares KFRA with existing methods on \textit{FGExpertBench} and traditional fine-grained image classification datasets, and Section~\ref{subsec:exp_analysis} provides further analysis of our method.

\subsection{Experimental Setting}
\label{subsec:exp_set}
\noindent \textbf{Implementation Details.}
KFRA is implemented as a tool-augmented reasoning framework coordinated by a Qwen3-A3B~\cite{yang2025qwen3} controller. 
The open-vocabulary detection module employs Grounding-DINO~\cite{liu2024grounding}, while the global-to-local focusing module is constructed upon VisionReasoner~\cite{liu2025visionreasoner}. 
For external retrieval, we adopt Google Lens for web-scale visual search and Wikipedia for textual knowledge reference. 
OseDiff~\cite{wu2024one} is integrated as a super-resolution enhancer to recover fine-grained details when critical cues are visually degraded. 
All experiments are conducted on a single NVIDIA H20 GPU under identical inference settings across all comparisons.

\begin{table}[!t]
\centering
\resizebox{0.98\columnwidth}{!}{
\begin{NiceTabular}{l|l|cccccc|c}
\toprule
& \textbf{Models}  & \textbf{Obj.} & \textbf{Attr.} & \textbf{Act.} & \textbf{Cnt.} & \textbf{Rsn.} & \textbf{Know.} & \textbf{Average} \\
\midrule
\Block{12-1}{\rotatebox{90}{\textbf{Open-Source}}}
& Phi-3.5-4B~\cite{abdin2024phi} & 36.45 & 46.94 & 51.02 & 47.33 & 56.00 & 45.76 & 47.47 \\
& Gemma3-it-4B~\cite{team2025gemma} & 45.81 & 51.02 & 55.10 & 51.33 & 45.33 & 54.00 & 50.43 \\
& LLaVA-OneVision-7B~\cite{li2024llava} & 31.03 & 36.73 & 48.98 & 40.67 & 34.67 & 44.27 & 39.39 \\
& LLaVA-Next-8B~\cite{li2024llavaN} & 31.53 & 51.02 & 57.14 & 44.00 & 36.67 & 42.24 & 43.77 \\
& InternVL3-8B~\cite{zhu2025internvl3} & 42.36 & 52.04 & 61.22 & 58.00 & 55.33 & 52.24 & 53.53 \\
& InternVL3.5-8B~\cite{wang2025internvl3} & 39.41 & 47.96 & 65.31 & 61.33 & 54.66 & 51.18 & 53.31 \\
& GLM-4.1V-9B~\cite{team2025glm} & 62.56 & 63.27 & 57.14 & 64.67 & 59.33 & 58.71 & 60.95 \\
& GLM-4.5V-12B~\cite{team2025glm} & 65.52 & 64.29 & 67.34 & 64.67 & 65.33 & 65.29 & 65.41 \\
& Qwen2.5-VL-7B~\cite{bai2025qwen2} & 33.50 & 42.86 & 63.27 & 46.67 & 34.00 & 41.53 & 48.64 \\
& Qwen2.5-VL-72B~\cite{bai2025qwen2} & 57.64 & 51.02 & 69.39 & 69.33 & 66.67 & 63.65 & 62.95 \\
& Qwen3-VL-8B~\cite{yang2025qwen3} & 51.72 & 63.27 & 75.51 & 53.33 & 56.67 & 56.71 & 59.54 \\
& Qwen3-VL-A22B~\cite{yang2025qwen3} & 64.53 & 61.22 & 73.47 & 66.67 & 67.33 & 66.24 & 66.58 \\
\midrule
\Block{5-1}{\rotatebox{90}{\textbf{Commercial}}}
& GLM-4V-Plus  & 47.29 & 45.92 & 63.27 & 64.67 & 58.00 & 48.35 & 54.58 \\
& Qwen-VL-Plus & 60.10 & 57.14 & 75.51 & 66.00 & 56.00 & 61.53 & 62.71 \\
& GPT-4o & 66.50 & 65.31 & 67.35 & 61.33 & 68.46 & 61.19 & 65.03 \\
& Doubao-1.5-Vision-Pro & 71.92 & 66.32 & 73.47 & 67.33 & 71.33 & 68.47 & 69.81 \\
& Gemini-2.5-Flash & 68.96 & 69.39 & 71.43 & 69.33 & 68.67 & 72.12 & 69.98 \\
\midrule
\Block{5-1}{\rotatebox{90}{\textbf{Agent-Based}}}
& VSA~\cite{zhang2024vision} & 31.68 & 44.90 & 51.02 & 31.33 & 38.67 & 37.81 & 39.24 \\
& MMSearch~\cite{jiang2024mmsearch} & 27.09 & 39.80 & 38.78 & 26.00 & 48.00 & 41.65 & 36.81 \\
& \Block[fill=mycolor]{1-1}{KFRA (Qwen2.5-VL-7B)} & \Block[fill=mycolor]{1-1}{68.47} & \Block[fill=mycolor]{1-1}{64.29} & \Block[fill=mycolor]{1-1}{75.51} & \Block[fill=mycolor]{1-1}{68.00} & \Block[fill=mycolor]{1-1}{64.67} & \Block[fill=mycolor]{1-1}{65.76} & \Block[fill=mycolor]{1-1}{67.78} \\
& \Block[fill=mycolor]{1-1}{KFRA (Qwen3-VL-8B)} & \Block[fill=mycolor]{1-1}{69.54} & \Block[fill=mycolor]{1-1}{72.45} & \Block[fill=mycolor]{1-1}{\textbf{83.67}} & \Block[fill=mycolor]{1-1}{65.33} & \Block[fill=mycolor]{1-1}{70.67} & \Block[fill=mycolor]{1-1}{69.41} & \Block[fill=mycolor]{1-1}{71.85} \\
& \Block[fill=mycolor]{1-1}{KFRA (GLM-4.5V-12B)} & \Block[fill=mycolor]{1-1}{\textbf{74.88}} & \Block[fill=mycolor]{1-1}{\textbf{74.49}} & \Block[fill=mycolor]{1-1}{77.55} & \Block[fill=mycolor]{1-1}{\textbf{71.33}} & \Block[fill=mycolor]{1-1}{\textbf{75.33}} & \Block[fill=mycolor]{1-1}{\textbf{75.29}} & \Block[fill=mycolor]{1-1}{\textbf{74.81}} \\
\bottomrule
\end{NiceTabular}
}
\caption{\textbf{Quantitative results (\%) on \textit{FGExpertBench.}} KFRA achieves strong and consistent performance across all dimensions, showing superior fine-grained perception and reasoning ability. \textbf{Bold} indicates the best performance in each column.}
\vspace{-1em}
\label{tab:table2}
\end{table}

\begin{figure*}[!t]
\centering
\includegraphics[width=0.98\linewidth]{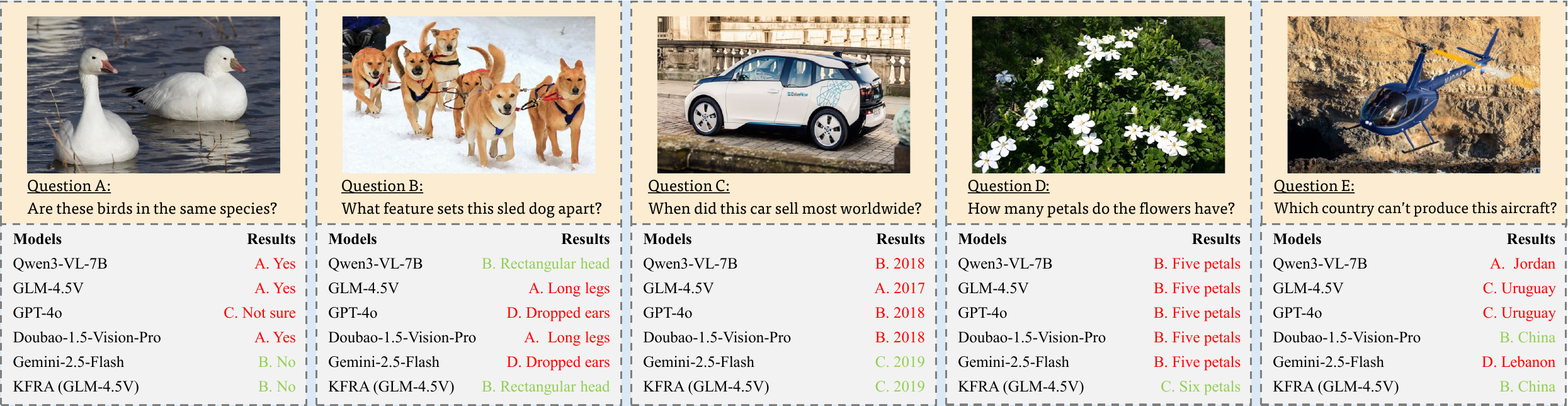}
\caption{\textbf{Qualitative results of all comparison methods on \textit{FGExpertBench}.} KFRA produces the most accurate answers across diverse fine-grained scenarios, highlighting the strong capability in visual reasoning and multimodal understanding.} 
\vspace{-1em}
\label{fig:figure6}
\end{figure*}

\noindent \textbf{Comparison Methods.} 
We compare KFRA against a comprehensive set of large multimodal models (LMMs) and recent agent-based frameworks. 
The evaluated LMMs cover both open-source and commercial systems to ensure fair and representative benchmarking. 
\textit{Open-source models} include Phi-3.5~\cite{abdin2024phi}, Gemma3-it~\cite{team2025gemma}, LLaVA-OneVision~\cite{li2024llava}, LLaVA-Next~\cite{li2024llavaN}, InternVL3 and InternVL3.5~\cite{zhu2025internvl3,wang2025internvl3}, GLM-4.1V and GLM-4.5V~\cite{team2025glm}, Qwen2.5-VL~\cite{bai2025qwen2}, and Qwen3-VL~\cite{yang2025qwen3}. 
\textit{Commercial models} include GLM-4V-Plus, Qwen-VL-Plus, GPT-4o, {Doubao-1.5-Vision-Pro}, and Gemini-2.5-Flash. For \textit{agent-based baselines}, we consider VSA~\cite{zhang2024vision} and MMSearch~\cite{jiang2024mmsearch}, which represent current paradigms for multimodal retrieval–reasoning agents. 
All models are evaluated using official checkpoints and default settings to ensure consistency across comparisons.

\noindent \textbf{Evaluation Datasets and Metrics.}
We primarily evaluate KFRA on the proposed \textit{FGExpertBench}, which measures reasoning depth and cross-task generalisation across six knowledge dimensions. 
To further assess the adaptability of KFRA to conventional recognition settings, we additionally test on six widely used fine-grained image classification (FGIC) datasets, including 
CUB-200-2011~\cite{wah2011caltech}, Stanford Cars~\cite{krause20133d}, Stanford Dogs~\cite{khosla2011novel}, Oxford 102 Flowers~\cite{nilsback2008automated}, FGVC-Aircraft~\cite{maji2013fine}, 
and Oxford-IIIT Pets~\cite{parkhi2012cats}. 
Following~\cite{he2025analyzing}, all datasets are reformulated into a unified question–answer format, where each question corresponds to a visual query with one correct answer among multiple candidates. 
Accuracy of model responses is reported as the evaluation metric, and the final score is computed as the average accuracy across all categories and dimensions.

\subsection{Comparison with Existing Methods}
\label{subsec:exp_compare}
\noindent\textbf{Quantitative Results.} 
Table~\ref{tab:table2} presents the overall results on \textit{FGExpertBench}. 
KFRA achieves consistent state-of-the-art performance across all six reasoning dimensions, substantially surpassing both open-source and commercial multimodal models. 
When built upon GLM-4.5V, it reaches the highest average accuracy of 74.81\%, outperforming Gemini-2.5-Flash by 4.83\%. 
When integrated with Qwen2.5-VL, KFRA delivers a remarkable 19.14\% improvement over its base model, demonstrating strong open-set generalisation and robustness across knowledge dimensions. 
Among agent-based frameworks, KFRA also shows clear superiority: VSA and MMSearch achieve only 39.24\% and 36.81\%, respectively, as their retrieval procedures lack fine-grained category alignment and fail to establish evidence-grounded reasoning. 
Note that KFRA particularly excels in \textit{Reasoning} and \textit{Knowledge} categories, underscoring its advantage in tasks that require explicit evidence construction and knowledge grounding.

\begin{figure*}[!t]
\centering
\includegraphics[width=0.98\linewidth]{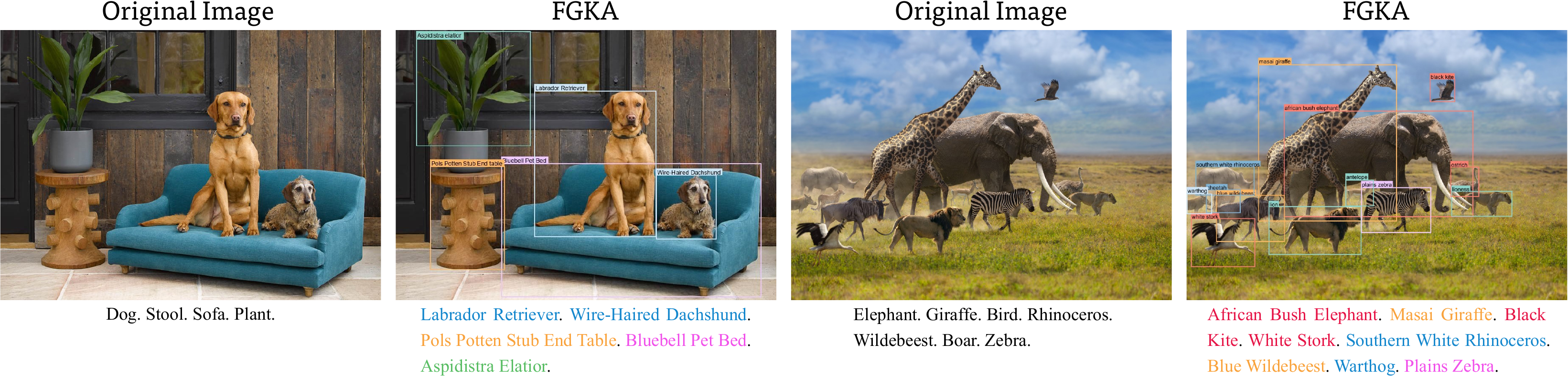}
\caption{\textbf{Visualisation of the ``Fine-Grained Anything'' Effect.} 
Starting from coarse detections guided by LMM-generated tags and Grounding-DINO, KFRA performs knowledge-driven reasoning, refining coarse predictions into precise fine-grained categories.}
\vspace{-1em}
\label{fig:figure7}
\end{figure*}

\begin{table}[!t]
\centering
\resizebox{0.98\columnwidth}{!}{
\begin{NiceTabular}{l|l|cccccc|c}
\toprule
& \textbf{Models} & \textbf{Bird} & \textbf{Car} & \textbf{Dog} & \textbf{Air.} & \textbf{Flwr.} & \textbf{Pet} & \textbf{Avg.} \\
\midrule
\Block{12-1}{\rotatebox{90}{\textbf{Open-Source}}}
& Phi-3.5-4B~\cite{abdin2024phi} & 40.66 & 44.31 & 41.52 & 30.57 & 48.02 & 61.87 & 44.49 \\
& Gemma3-it-4B~\cite{team2025gemma} & 50.05 & 91.07 & 66.66 & 66.82 & 81.25 & 86.35 & 73.70 \\
& LLaVA-OneVision-7B~\cite{li2024llava} & 34.14 & 61.77 & 51.99 & 53.56 & 51.21 & 62.06 & 52.46 \\
& LLaVA-Next-8B~\cite{li2024llavaN} & 31.58 & 37.56 & 29.27 & 18.36 & 40.15 & 46.09 & 33.84 \\
& InternVL3-8B~\cite{zhu2025internvl3} & 51.28 & 70.80 & 58.47 & 57.40 & 65.47 & 80.49 & 63.99 \\
& InternVL3.5-8B~\cite{wang2025internvl3} & 52.50 & 71.38 & 52.27 & 58.78 & 58.74 & 78.82 & 62.13 \\
& GLM-4.1V-9B~\cite{team2025glm} & 65.96 & 72.84 & 64.86 & 65.74 & 85.20 & 91.31 & 74.32 \\
& GLM-4.5V-12B~\cite{team2025glm} & 83.53 & 94.79 & 85.94 & 77.23 & 90.68 & 95.34 & 87.92 \\
& Qwen2.5-VL-7B~\cite{bai2025qwen2} & 41.08 & 68.05 & 56.43 & 47.19 & 74.52 & 81.68 & 61.49 \\
& Qwen2.5-VL-72B~\cite{bai2025qwen2} & 74.09 & 81.84 & 69.24 & 61.24 & 87.14 & 86.89 & 76.74 \\
& Qwen3-VL-8B~\cite{yang2025qwen3} & 38.95 & 67.52 & 51.79 & 45.00 & 80.87 & 76.70 & 60.14 \\
& Qwen3-VL-A22B~\cite{yang2025qwen3} & 73.44 & 84.57 & 71.29 & 70.33 & 89.69 & 86.05 & 79.23 \\
\midrule
\Block{3-1}{\rotatebox{90}{\textbf{Com.}}}
& GPT-4o & 81.36 & 93.21 & 83.30 & \textbf{85.06} & 93.17 & 91.77 & 87.98 \\
& Doubao-1.5-Vision-Pro & 81.12 & 86.90 & 79.97 & 73.12 & 92.01 & 92.56 & 84.28 \\
& Gemini-2.5-Flash & 84.41 & 92.89 & 84.42 & 83.92 & 93.72 & 95.23 & 89.10 \\
\midrule
\Block{4-1}{\rotatebox{90}{\textbf{SOTA}}}
& Finedefics-8B~\cite{he2025analyzing} & 57.61 & 84.67 & 72.86 & 63.82 & 89.88 & 92.18 & 76.84 \\
& DeepPerception-7B~\cite{ma2025deepperception} & 67.86 & 91.75 & 78.29 & 75.31 & 93.06 & 93.00 & 83.21 \\
& \Block[fill=mycolor]{1-1}{KFRA (Qwen2.5-VL-7B)} & \Block[fill=mycolor]{1-1}{83.28} & \Block[fill=mycolor]{1-1}{92.41} & \Block[fill=mycolor]{1-1}{79.24} & \Block[fill=mycolor]{1-1}{80.56} & \Block[fill=mycolor]{1-1}{83.23} & \Block[fill=mycolor]{1-1}{91.88} & \Block[fill=mycolor]{1-1}{85.10} \\
& \Block[fill=mycolor]{1-1}{KFRA (GLM-4.5V-12B)} & \Block[fill=mycolor]{1-1}{\textbf{86.52}} & \Block[fill=mycolor]{1-1}{\textbf{95.85}} & \Block[fill=mycolor]{1-1}{\textbf{87.66}} & \Block[fill=mycolor]{1-1}{81.04} & \Block[fill=mycolor]{1-1}{\textbf{93.83}} & \Block[fill=mycolor]{1-1}{\textbf{96.51}} & \Block[fill=mycolor]{1-1}{\textbf{90.24}} \\
\bottomrule
\end{NiceTabular}
}
\caption{\textbf{Experimental results (\%) on six conventional FGIC datasets.} KFRA achieves the competitive performance across all datasets. \textbf{Bold} {indicates} the best performance in each column.}
\vspace{-1.5em}
\label{tab:table3}
\end{table}

\noindent\textbf{Qualitative Results.}
Figure~\ref{fig:figure6} provides qualitative comparisons illustrating how KFRA performs reasoning through explicit evidence construction. 
Across diverse fine-grained scenarios, KFRA produces correct and interpretable answers where existing large multimodal models exhibit semantic confusion or local bias. 
In \textit{Question A}, two visually similar species—Snow Goose and Ross Goose—are correctly distinguished by KFRA through localisation of diagnostic cues around the beak and wing bar, while most other models collapse them into a single class. 
In \textit{Question B}, KFRA identifies a Chinook breed by grounding textual traits such as ``black muzzle'' and ``arched neck'', which competing models misinterpret due to missing fine-scale attention. 
{\textit{Questions C} and \textit{D}} highlight structural reasoning: KFRA recognises the BMW i3 model year and correctly infers the petal structure of Gardenia jasminoides, avoiding the common bias toward more salient but irrelevant features. 
Finally, in \textit{Question E}, the agent links a Robinson R44 helicopter to its correct origin, demonstrating factual reasoning via retrieved external knowledge. 
These examples collectively show that KFRA’s closed reasoning loop: linking retrieval, grounding, and verification, which enables factual, spatially grounded, and interpretable fine-grained understanding that generalises beyond pattern recognition.

\subsection{Further Analysis}
\label{subsec:exp_analysis}
\noindent\textbf{Comparison on FGIC Datasets.}
To validate the generalisation capability of KFRA, we evaluate its performance on six conventional fine-grained image classification (FGIC) benchmarks, as reported in Table~\ref{tab:table3}. 
Despite being trained without any task-specific supervision, KFRA achieves highly competitive results across all datasets. 
When built upon GLM-4.5V, it attains the best average accuracy of 90.24\%, surpassing Gemini-2.5-Flash (89.10\%) and DeepPerception-7B (83.21\%). 
Even with a lighter backbone such as Qwen2.5-VL-7B, KFRA still reaches 85.10\%, marking a 23.61\% improvement over the base model. 
This substantial gain demonstrates that the proposed closed reasoning loop not only enhances interpretability but also improves recognition robustness in traditional classification settings. 

\noindent\textbf{Fine-Grained Anything.} 
To assess the broader applicability of KFRA, we extend it to open-world scenarios, demonstrating its potential for ``fine-grained anything'' reasoning. 
Specifically, GLM-4.5V first analyses the image context to generate descriptive tags, which are then used as prompts for Grounding-DINO to guide spatial localisation. 
The detected regions are subsequently refined by KFRA for category-level reasoning, as shown in Figure~\ref{fig:figure7}. 
KFRA accurately localises and distinguishes fine-grained categories that models fail to recognise, revealing that its closed reasoning loop extends beyond question answering. 
By coupling perception with knowledge-driven inference, KFRA transforms generic open-vocabulary detectors into fine-grained experts capable of precise recognition in complex open-world settings.

\noindent\textbf{Ablation Study.}
To evaluate the role of each component, we conduct a removal ablation on \textit{FGExpertBench} using Qwen2.5-VL-7B as the backbone and the standalone model (48.64\%) as baseline. 
Table~\ref{tab:table4} shows that accuracy gradually increases as more tools are enabled.
When only perception modules (VS, OD, GF, SR) are used, the accuracy rises slightly to 49.56\%, indicating limited improvement without factual grounding. 
Introducing the knowledge reference (KR) brings steady gains, reaching 50.21\% and confirming the importance of external knowledge. 
With all modules activated, KFRA achieves 67.78\%, a +19.14\% overall improvement. 
KR contributes the most, while VS and OD ensure reliable hypotheses, and GF with SR enhance spatial alignment and fine-detail verification.
These results demonstrate that retrieval and grounding act synergistically, forming the closed reasoning loop that transforms large multimodal models into evidence-driven reasoning agents.

\begin{table}[t]
\centering
\resizebox{0.4\textwidth}{!}{
\fontsize{7pt}{9pt}\selectfont
\begin{tabular}{ccccc|cc}
\toprule
\textbf{KR} & \textbf{VS} & \textbf{OD} & \textbf{GF} & \textbf{SR} & \textbf{Acc. \%} & \textbf{Imp. \%}\\
\midrule
&  &  &  &  & 48.64 & -\\
& \checkmark & \checkmark & \checkmark & \checkmark & 49.56 & +0.92\\
\checkmark &  & \checkmark & \checkmark & \checkmark & 50.21 & +1.57\\
\checkmark & \checkmark &  & \checkmark & \checkmark & 52.83 & +4.19\\
\checkmark & \checkmark & \checkmark &  & \checkmark & 57.15 & +8.51\\
\checkmark & \checkmark & \checkmark & \checkmark &  & 63.42 & +14.78\\
\checkmark & \checkmark & \checkmark & \checkmark & \checkmark & \textbf{67.78} & \textbf{+19.14}\\
\bottomrule
\end{tabular}
}
\caption{\textbf{Ablation study of tools within KFRA on \textit{FGExpertBench}.} \#Acc. denotes the average accuracy. \#Imp. denotes the improvement above baseline. \textbf{Bold} {indicates} the best performance.}
\vspace{-1.4em}
\label{tab:table4}
\end{table}

\section{Limitations}
Although KFRA achieves strong generalisation and interpretability, it still faces several limitations. 
The framework relies on external APIs for retrieval and knowledge access, which may cause latency or errors in retrieved information. 
Its multi-step design increases computational cost compared with single-pass models. 
In addition, \textit{FGExpertBench} currently has limited domain coverage, and broader extensions would further validate generality.

\section{Conclusion}
We introduced the \textbf{Knowledge-Augmented Fine-Grained Reasoning Agent (KFRA)}, a framework that transforms recognition into reasoning through a closed loop of perception, grounding, and inference. 
KFRA converts retrieved knowledge into verifiable spatial evidence, enabling factual and interpretable reasoning across open-set scenarios. 
We also proposed \textit{FGExpertBench} to evaluate reasoning depth and generalisation. 
Experiments confirm that KFRA advances fine-grained understanding beyond classification toward expert-level reasoning.

\noindent\textbf{Acknowledgements.} This work was supported by the National Natural Science Foundation of China (Grant 62406171, 62225601, U23B2052, 62306031), in part by the Beijing Natural Science Foundation Project No. L242025, in part by the Guizhou Province Science and Technology Plan Project (No. QianKeHeZhongDa [2025]031), and in part by the Fundamental Research Funds for the Beijing University of Posts and Telecommunications under Grant 2025AI4S15.

{
    \small
    \bibliographystyle{ieeenat_fullname}
    \bibliography{main}
}

\end{document}